\documentclass[times,twocolumn,final,authoryear]{elsarticle}

\usepackage{prletters}
\usepackage{framed,multirow}

\usepackage{amssymb}
\usepackage{latexsym}

\usepackage{url}
\usepackage{xcolor}
\definecolor{newcolor}{rgb}{.8,.349,.1}

 \usepackage[T1]{fontenc} 
\usepackage[utf8]{inputenc} 
\usepackage[portuguese]{babel} 

\usepackage{amsmath,bm}
\usepackage{array}
\usepackage[inline]{enumitem}
\usepackage{tikz}
\usetikzlibrary{automata,positioning,patterns}
\usepackage{graphicx}
\usepackage{xcolor,pifont} \newcommand*\colourcheck[1]{\expandafter\newcommand\csname #1check\endcsname{\textcolor{#1}{\ding{52}}}}
\usepackage[compact]{titlesec}
\titlespacing{\section}{3pt}{2.5ex}{1ex}
\titlespacing{\subsection}{2pt}{1.5ex}{0.3ex}
\titlespacing{\subsubsection}{1pt}{.3ex}{0.2ex}
\newcolumntype{P}[1]{>{\centering\arraybackslash}p{#1}}
\newcolumntype{M}[1]{>{\centering\arraybackslash}m{#1}}

\usepackage{dblfloatfix} 

\journal{Pattern Recognition Letters}

\begin{document}

\begin{frontmatter}
\title{Person Re-identification: Implicitly Defining the Receptive Fields of Deep Learning Classification Frameworks}

\author[1]{Ehsan \snm{Yaghoubi}\corref{cor1}} 
\cortext[cor1]{Corresponding author}  
\ead{Ehsan.yaghoubi@ubi.pt}
\author[2]{Diana \snm{Borza}}
\author[3]{S V Aruna \snm{Kumar}}
\author[1]{Hugo \snm{Proença}}

\address[1]{IT: Instituto de Telecomunicações, University of Beira Interior, Covilhã, Portugal}
\address[2]{Technical University of Cluj-Napoca, Romania}
\address[3]{University of Beira Interior, Covilhã, Portugal}


\begin{abstract}
The \emph{receptive fields} of deep learning classification models determine the regions of the input data that have the most significance for providing correct decisions. The primary way to learn such receptive fields is to train the models upon masked data, which helps the networks to ignore any unwanted regions, but has two major drawbacks: 1) it often yields edge-sensitive decision processes; and 2) augments the computational cost of the inference phase considerably. This paper describes a solution for implicitly driving the inference of the networks' receptive fields, by creating synthetic learning data composed of interchanged segments that should be \emph{apriori} important/irrelevant for the network decision. In practice, we use a segmentation module to distinguish between the foreground (important)/background (irrelevant) parts of each learning instance, and randomly swap segments between image pairs, while keeping the class label exclusively consistent with the label of the deemed important segments. This strategy typically drives the networks to early convergence and appropriate solutions, where the identity and clutter descriptions are not correlated. Moreover, this data augmentation solution has various interesting properties: 1) it is parameter-free;  2) it fully preserves the label information; and, 3) it is compatible with the typical data augmentation techniques.  In the empirical validation, we considered the person re-identification problem and evaluated the effectiveness of the proposed solution in the well-known \emph{Richly Annotated Pedestrian} (RAP) dataset for two different settings (\emph{upper-body} and \emph{full-bod}y), observing highly competitive results over the state-of-the-art. Under a reproducible research paradigm, both the code and the empirical evaluation protocol are available at \url{https://github.com/Ehsan-Yaghoubi/reid-strong-baseline}.



\end{abstract}

\end{frontmatter}











\section{Introduction}
\label{sec:intro}
Person re-identification (re-id) refers to the cross-camera retrieval task, in which a query from a target subject is used to retrieve identities from a gallery set. This process is tied to many difficulties, such as variations in human poses, illumination, partial occlusions, and cluttered backgrounds. The primary way to address these challenges is to provide large-scale \textit{labeled} learning data (which are not only hard to collect but particularly costly to annotate) and expect that the deep model learn the critical parts of the input data autonomously. This strategy is supposed to work for any problem, upon the existence of enough learning data, which might correspond to millions of learning instances in case of hard problems. 

To skim the costly annotation step, various works have proposed to augment the learning data using different techniques \citep{shorten2019survey}. They either use the available data to synthesize new images or generate new images by sampling from the learned distribution. In both cases, the main objective is to increase the quantity of data, without assisting the model to find the relevant regions of the input, so that often the networks find spurious patterns in the background regions that --yet-- are matched with the ground truth labels. This kind of techniques shows positive effects in several applications; for example, \citep{dvornik2019importance} proposes an object detection model, in which the objects are cut out from their original background and pasted to other scenes (e.g., a plane is pasted between different sky images). On the contrary, in the pedestrian attribute recognition and re-identification problems, the background clutter is known as a primary obstacle to the reliability of the inferred models. 

\begin{figure}
    \centering
    \includegraphics[width=\linewidth]{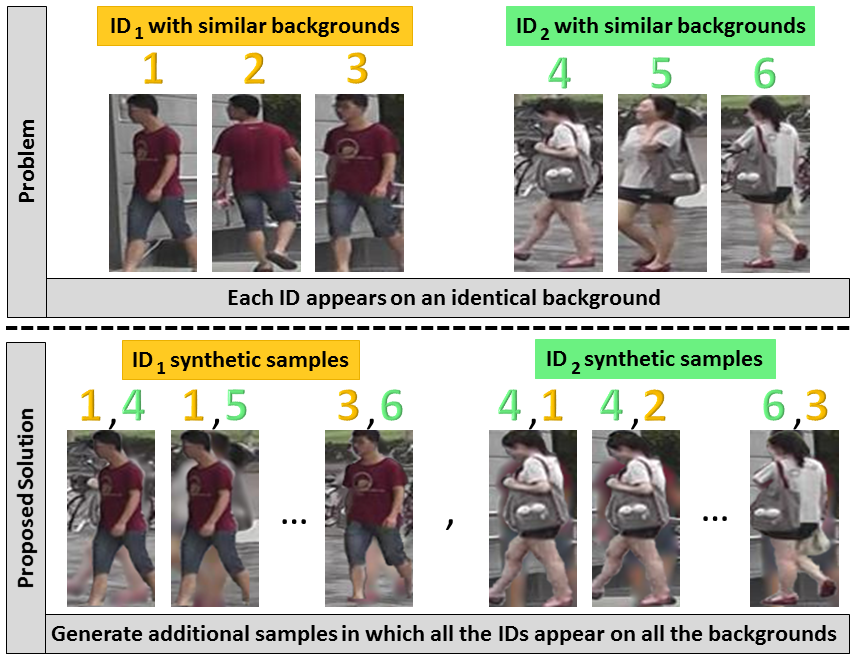}
    \caption{The main challenge addressed in this paper: during the learning phase, if the model sees all samples of one ID in a single scene, the final feature representation of that subject might be entangled with spurious (background) features. By creating synthetic samples with multiple backgrounds, we implicitly \emph{guide} the network to focus on the deemed important features.}
    \label{fig:challenge}
\end{figure}

Holistic CNN-based re-id models extract global features, regardless of any critical regions in the input data, and typically fail when the background covers most of the input. In particular, when dealing with limited amounts of learning data, three problems emerge: 1) holistic methods may not find the foreground regions automatically; 2) part-based methods \citep{varior2016siamese}, \citep{li2017learning} typically fail to detect the appropriate critical regions; and 3) attention-based models (e.g., \citep{xu2018attention} and \citep{zhao2017deeply}) face difficulties in case that multiple persons appear in a single bounding box.
As an attempt to reduce the classification bias due to the background clutter (caused by inaccurate person detection or crowded scenes), \citep{zheng2018pedestrian} proposes an alignment method to refine the bounding boxes, while \citep{zhang2017alignedreid} uses a local feature matching technique. As illustrated in Fig. \ref{fig:challenge}, although the alignment-based re-id approaches reduce the amounts of cluttered in the learning data, the networks still typically suffer from the remaining background features, particularly if some of the IDs always appear in the same scene (background). 

To address the above-described problems,  this paper introduces a receptive field implicit definition method based on data augmentation that could be applied to the existing re-id methods as a complementary step. The proposed solution is: 1) mask-free for the \emph{test} phase, i.e., it does not require any additional explicit segmentation in test time; and 2) contributes to foreground-focused decisions in the inference phase. The main idea is to generate synthetic data composed of interleaved segments from the original learning set, while using class information only from specific segments. During the learning phase, the newly generated samples feed the network, keeping their label exclusively consistent with the identity from where the region-of-interest was cropped. Hence, as the model receives images of each identity with inconsistent unwanted areas (e.g., background), it naturally pays the most attention to the regions consistent with ground truth labels. We observed that this pre-processing method is equivalent to only learn from the effective receptive fields and ignore the destructive regions. During the test phase, samples are provided without any mask, and the network naturally disregards the detrimental information, which is the insight for the observed improvements in performance.
 
In particular, when compared to \citep{Augmentation} and \citep{zhong2020random}, this paper can be seen as a data augmentation technique with several singularities: 1) we not only enlarge the learning data but also implicitly provide the inference model with an attentional decision-making skill, contributing to \textit{ignore} irrelevant image features during the test phase; 2) we generate highly representative samples, making it possible to use our solution along with other data augmentation methods; and 3) our solution allows the on-the-fly data generation, which makes it efficient and easy to be implemented beside the common data augmentation techniques. Our evaluation results point for consistent improvements in performance when using our solution over the state-of-the-art person re-id method.

\section{Related Work}
\label{sec:Related_work}

\textbf{\textit{Data Augmentation}}.
Data augmentation targets the root cause of the over-fitting problem by generating new data samples and preserving their ground truth labels. \emph{Geometrical transformation}  (scaling, rotations, flipping, etc.), \emph{color alteration} (contrast, brightness, hue), \emph{image manipulation} (random erasing \citep{zhong2020random}, kernel filters, image mixing \citep{Augmentation}), and \emph{deep learning approaches} (neural style transfer, generative adversarial networks) \citep{shorten2019survey} are the common augmentation techniques. 

Recently, various methods have been proposed for image synthesizing and data augmentation \citep{shorten2019survey}. For example, \citep{Augmentation} generates $n^2$ samples from an $n$-sized dataset by using a sample pairing method, in which a random couple of images are overlaid based on the average intensity values of their pixels. \citep{zhong2020random} presents a \emph{random erasing} data augmentation strategy that inflates the learning data by randomly selecting rectangular regions and changing their pixels values. As an attempt to robust the model against occlusions, increasing the volume of the learning data turned the concept of \emph{random erasing} into a popular data augmentation technique. \citep{dvornik2019importance} addressed the subject of object detection, in which the background has helpful features for detecting the objects; therefore, authors developed a context-estimator network that places the instances (i.e., cut out objects) with meaningful sizes on the relevant backgrounds.


\textbf{\textit{Person Re-ID}}.
In general, early person re-id works studied either the descriptors to extract more robust feature representations or metric-based methods to handle the distance between the inter-class and intra-class samples \citep{bedagkar2014survey}. However, recent re-id studies are mostly based on deep learning neural networks that can be classified into three branches \citep{wu2019deep}: Convolutional Neural Network (CNN), CNN-Recurrent neural network, and Generative Adversarial Network (GAN). 

Among the CNN and CNN-RNN methods, those based on attention mechanisms follow a similar objective to what we pursue in this paper; i.e., they ignore background features by developing attention modules in the backbone feature extractor. Attention mechanism may be developed for either single-shot or multi-shot (video) \citep{chen2019spatial}, \citep{zhang2020ordered}, \citep{cheng2020scale} scenarios, both of them aim to learn a distinctive feature representation that focuses on the critical regions of the data. To this end, \citep{yang2019attention} use the body-joint coordinates to remove the extra background and divide the image into several horizontal pieces to be processed by separate CNN branches. \citep{xu2018attention} and \citep{zhao2017deeply} propose a body-part detector to re-identify the probe person with matching the bounding boxes of each body-part, while \citep{zhou2020mask} uses the masked out body-parts to ignore the background features in the matching process. In contrast to these works that explicitly implement the attentional process in the structure of the neural network \citep{denil2012learning}, we provide an attentional control ability based on receptive field augmentation detailed in section \ref{sec:Proposed_method}. Therefore, in some terms, our work is similar to the GAN-based re-id techniques, which usually aim to either increase the quantity of the data \citep{zheng2017unlabeled} or present novel poses of the existing identities \citep{liu2018pose}, \citep{borgia2019gan} or transfer the camera style \citep{lin2020unsupervised}, \citep{wei2018person}. Although GAN-based works present novel features for each individual, they generate some destructive features that are originated from the new backgrounds. Furthermore, these works ignore to handle the problem of co-appearance of multiple identities in one shot.

\section{Proposed Method}
\label{sec:Proposed_method}
\begin{figure*}
\centering
\scriptsize
\begin{tikzpicture}
\colourcheck{green}
\node[anchor=south west] (image) at (-10,0) {\includegraphics[width=11.5cm]{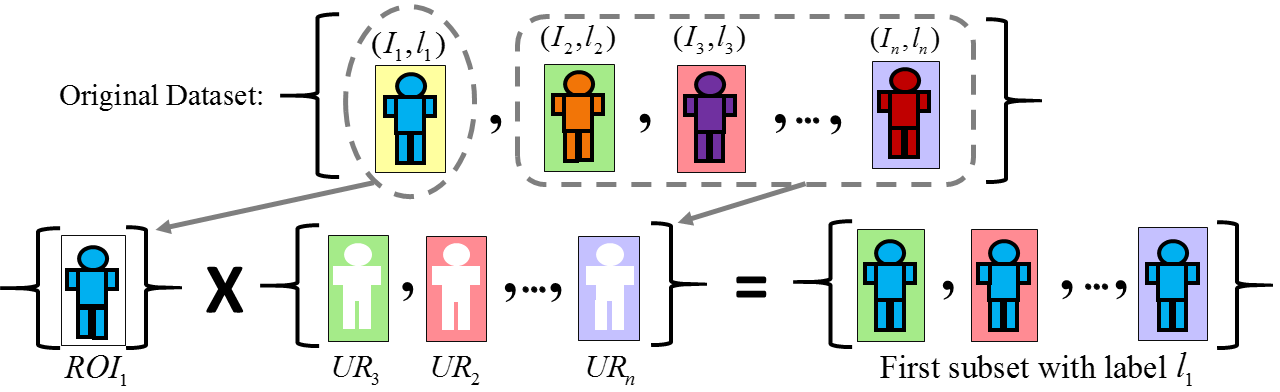}};

\node [text width=6cm, align=left][anchor=west] at (2 , 3.4) {\textbf{\underline{ADVANTAGES}}};

\node [text width=6cm, align=left][anchor=west] at (2 , 2.5) {\greencheck Attention on the Region of Interest (ROI) during the test phase without using mask segmentation};

\node[text width=6cm, align=left][anchor=west] at (2, 1.8) {\greencheck Inflating the learning set from $n$ samples to $n^2$ samples};

\node[text width=6cm, align=left][anchor=west] at (2, 1.3) {\greencheck Complementary to common data augmentation methods};

\node[text width=6cm, align=left][anchor=west] at (2, 0.3) {\greencheck Parameter learning free technique};

\node[text width=6cm, align=left][anchor=west] at (2, 0.8) {\greencheck Generating highly representative samples};
\end{tikzpicture}
    
    \caption{The proposed full-body attentional data augmentation (best viewed in color). Blue, orange, purple, and red denote the samples $1$, $2$, $3$, and $N$, respectively. The pale-yellow, green, pink, and purple colors represent their cluttered (background) regions, which should be irrelevant for the inference process. Therefore, all the synthetic images labeled as $1$ share the blue body region but have different backgrounds, which provides a strong cue for the network to disregard such segments from the decision process.}%
    \label{fig:attentional mechanism} %
    \vspace{-0.4 cm}
\end{figure*}

Figure \ref{fig:attentional mechanism} provides an overview of the proposed image synthesis method, in this case, considering the full-body as the region of interest (ROI). We show the first synthesize subset, in which the new samples comprise of the $ROI$ of the $1^{st}$ sample and the background of the other samples. It worth mentioning that we have also evaluated the method for focusing on the upper-body region, such that the generated images are composed of the upper-body of the $1^{st}$ sample and remaining segments (background and lower-body regions) of the other images while keeping the label of the $1^{st}$ sample.

\subsection{Implicit Definition of Receptive Fields}
\label{subsec:Augmentation_Based_Attention}

As an intrinsic behavior of CNNs, in the learning phase, the network extracts a set of essential features in accordance with the image annotations. However, extracting relevant and compressed features is an ongoing challenge, especially when the background\footnote{The terms \emph{(unwanted region/region-of-interest)}, \emph{(undesired/desired) boundaries}, \emph{(background/foreground) areas}, and \emph{(unwanted/wanted) areas} refer to the data segments that are deemed to be irrelevant/relevant to the ground truth label. For example, in a hair color recognition problem, the region-of-interest is the hair area, which can be defined by a binary mask} changes with person ID. Intuitively, when a person's identity appears with an identical background, some background features are entangled with the useful foreground features and reduce the inference performance. However, if the network sees one person with different backgrounds, it can automatically discriminate between the relevant regions of the image and the ground truth labels.\textbf{ Therefore, to help the \emph{inference model} automatically distinguishes between the unwanted features and foreground features, in the \emph{learning phase}, we repeatedly feed synthetically generated, fake images to the network that has been composed of two components:} 
\begin{enumerate*} [label=\roman*)]
\item critical parts of the current input image that describe the ground truth labels (i.e., person's identity), and we would like to have an attention on them, and \item parts of the other real samples that intuitively are uncorrelated with the current identity (i.e., background and body parts that we would like the network to ignore them).
\end{enumerate*} 
\textbf{Thus, the model looks through each region of interest, juxtaposed with different unwanted regions --of all the images-- enabling the network to learn where to look at in the image and ignores the parts that are changing arbitrarily and are not correlated with ground truth labels.}
Consequently, during the test phase, the model explores the region of interest and discards the features of unwanted regions that have been trained for.

Formally, let $\bm{I}_i$, represent the i$^{th}$ image in the learning set, $\bm{l}_i$, its ground truth label and $\bm{M}_j$ the corresponding binary mask, discriminating the foreground/background regions. If $\bm{ROI}_.$ shows the region of interest and $\bm{UR}_.$ refers to the unwanted region, then, the goal is to synthesis the artificial sample  
$\bm{S}_{i\neg j}$ with the label $\bm{l}_i$: 

\begin{equation}
  \bm{S}_{i\neg j}(x, y) =  \bm{ROI}_i  \cup \bm{UR}_j,
\end{equation} 
where $\bm{ROI}_.= \bm{I}_{.}(x, y)$ such that $\bm{M}_{.}(x, y) = 1$, $\bm{UR}_.= \bm{I}_{.}(x, y)$ such that $\bm{M}_{.}(x, y) = 0$, and $(x,y)$ are the coordinates of the pixels.


Therefore, for an $n$-sized dataset, the \emph{maximum} number of generated images is equal to $n^2-n$. However, to avoid losing the expressiveness of the generated samples, we consider several constraints. Hence, a combination of the common data transformations (e.g., flipping, cropping, blurring) can be used along with our method. Obviously, since we utilize the ground truth masks, our technique should be done in the first place before any other augmentation transformation.

\subsection{Synthetic Image Generation}
\label{subsec:Generate_Synthesis_Images}

To ensure that the synthetically generated images have a natural aspect, we impose the following constraints: 

\textbf{\textit{Size and shape constraint}}. We avoid combining images with significant differences in their aspect ratios of the ROIs to circumvent the unrealistic stretching of the replaced content or large discontinuities between the body parts in the generated images. To this end, the ratio between the foreground areas defined by masks $\bm{M}_j$ and $\bm{M}_i$ should be less than the threshold $\bm{T}_s$ (we considered $\bm{T}_s=0.8$ in our experiments). Let $\bm{A}_.$ be the area of the foreground region (i.e., mask $\bm{M}_.$):
\begin{equation}
    \bm{A}_. = \sum_{x=0}^{w}\sum_{y=0}^{h} M_j(x, y),
\end{equation}

where $w$ and $h$ are the width and height of the image, respectively. This constraint translates to ${\min({\bm{A}_i, \bm{A}_j}})/{\max({\bm{A}_i, \bm{A}_j}}) < \bm{T}_s$. Moreover, to ensure the shape similarity, we calculate the Intersection over Union metric (IoU) for masks $\bm{M}_i$ and $\bm{M}_j$. 
\begin{equation}
    IoU(\bm{M}_i, \bm{M}_j) = \frac{\bm{M}_i \cap \bm{M}_j}{ \bm{M}_i \cup \bm{M}_j}.
\end{equation}
    
By preserving the aspect ratio, the rectangular masks are then cropped and resized to normalize their dimensions. Afterward, if the $IoU(\bm{M}_i, \bm{M}_j)$ is lower than a threshold $\bm{T}_i$, we consider those images for the merging process.
    
    \textbf{\textit{Smoothness constraint}}.  The transition between the source image and the replaced content should be as smooth as possible to prevent from inserting any strong edges. As a challenge, $\bm{M}_i$ and the body silhouette of the \emph{j}-th person do not match perfectly. To overcome this issue, we enlarge the mask $\bm{M}_j$ by using the morphological dilation operator with a $7 \times 7$ kernel: $\bm{M}_d = \bm{M}_j \oplus K_{7\times7}$. Next, to guarantee the continuity between the background and the newly added content, we use the image in-painting technique in \citep{telea2004image} to remove the undesired area from the source image, as it has been dictated by the enlarged mask $\bm{M}_d$.

\begin{figure*}%
\raggedleft
\begin{tikzpicture}
\colourcheck{green}

\node[anchor=center] (p1_0) at (0,0) {\includegraphics[width=\linewidth]{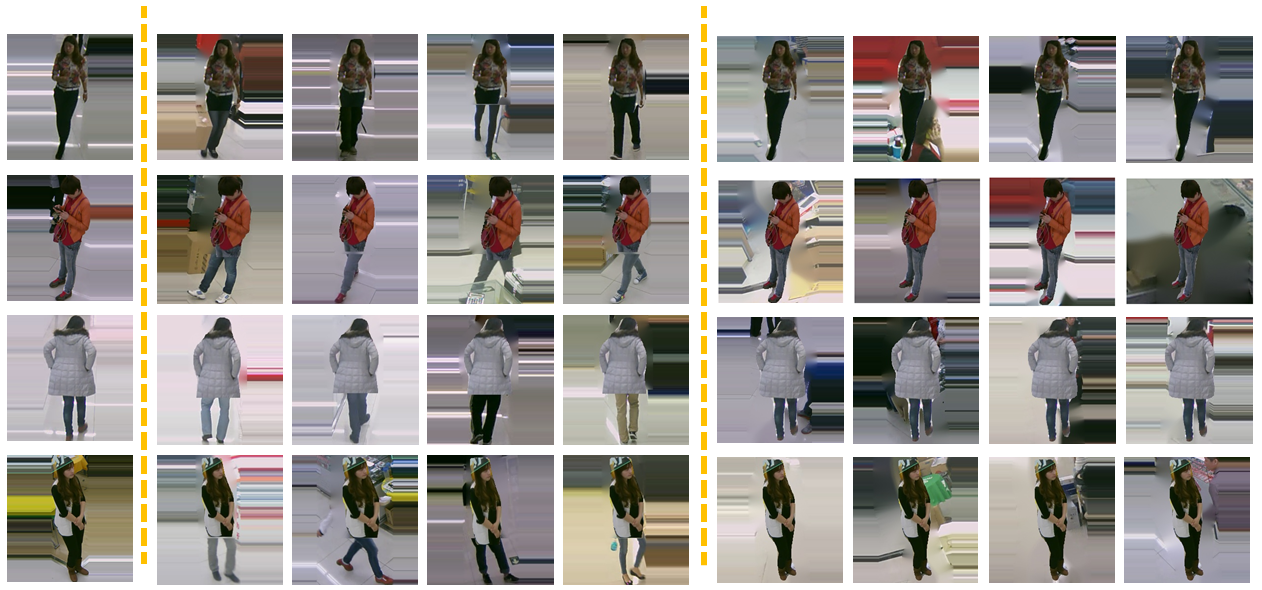}};

\node[align=center] at (-8.3,4) {\footnotesize Original};
\node[align=center] at (5,4) {\footnotesize Synthetic samples (Full body receptive field)};
\node[align=center] at (-3,4) {\footnotesize Synthetic samples (Upper body receptive field)};

\end{tikzpicture}
\caption{Examples of synthetic data generated for upper-body (center columns) and full-body (rightmost columns) receptive fields. The leftmost column shows the original images. Additional examples are provided at \emph{https://github.com/Ehsan-Yaghoubi/reid-strong-baseline}.}%
\label{fig:example}%
\vspace{-0.4 cm}
\end{figure*}

\textbf{\textit{Viewpoint constraint}\footnote{This step is applicable for body-part data augmentation (see Fig. \ref{fig:example}).}}. Body pose is another decisive factor for generating semantically natural images. Thereby, we only combine images with the same viewpoint annotations. For instance, suppose that we are interested in building attention on the human upper-body; therefore, we should avoid generating images composed of anterior upper-body of the \emph{i}-th person and posterior lower-body (and background) of the \emph{j}-th person. One can apply Alphapose \citep{fang2017rmpe} on any pedestrian dataset to estimate the body poses and then, uses \citep{maaten2008visualizing} to create clusters of poses as the viewpoint label. However, since the RAP dataset annotations include the pose information, we skipped this step in our experiments.

Figure \ref{fig:example} shows some examples generated by our technique, providing attention to the upper-body or full-body region. When enabling the attention of the upper-body region, fake samples are different in the human lower body and the environment, while they resemble each other in the person's upper body and identity label. By selecting the full-body as the ROI, the generated images will be composed of similar body silhouettes with different surroundings.

\section{Implementation Details}
\label{sec:Implementation_Details}
We based our method on the baseline \citep{strong_baseline} and selected a similar model architecture, parameter settings, and optimizer. In this baseline\footnote{https://github.com/michuanhaohao/reid-strong-baseline}, authors resized images on-the-fly into $128 \times 128$ pixels. As the RAP images vary in resolution (from $33 \times 81$ to $415 \times 583$), to avoid any data deformation, we first mapped the images to a squared shape, using a \emph{replication} technique, in which the row or column at the very edge of the original image is replicated to the extra border of the image.

As the RAP dataset does not provide human body segmentation annotations, we first fed the images to Mask-R-CNN model \citep{he2017mask}, which allowed to obtain the human segmentation binary masks. Next, as described in subsection \ref{subsec:Generate_Synthesis_Images}, we generated the synthetic images, using the default parameter settings described in the official project web page\footnote{https://github.com/matterport/Mask\_RCNN} and inferred the person masks without modifying the model weights.

 To perform the re-id process, we followed the instructions of \citep{li2018richly} and split the dataset into train ($1,295$ ids), query ($1,294$ ids), and gallery ($1,294$ ids) sets, including respectively $13,101$, $7173$, and $13,392$  samples from $23$ different cameras. 
 
\section{Experiments and Discussion}
\label{sec:Experiments}

\subsection{Datasets}
\label{subsec:Data}
 The \emph{Richly Annotated Pedestrian} (RAP) benchmark \citep{li2018richly} is the largest well-known pedestrian dataset composing of around $85,000$ samples from which $41,585$ images have been elected manually for identity annotation. The RAP re-id set includes $26,638$ images of $2,589$ identities and $14,947$ samples as distractors that have been collected from $23$ cameras in a shopping mall. The provided human bounding boxes have different resolutions ranging from $33 \times 81$ to $415 \times 583$. In addition to human attributes, the RAP dataset is annotated for camera angle, body-part position, and occlusions. 
 
\subsection{Baseline}
\label{subsec:Baseline}
A  recent work by Facebook AI \citep{musgrave2020metric} mentions that upgrading factors such as the learning method (e.g., \citep{roth2019mic}, \citep{kim2018attention}), network architecture (e.g., ResNet, GoogleNet, BN-Inception), loss function (e.g., embedding losses \citep{cakir2019deep}, \citep{wang2019multi} and classification losses \citep{wang2018cosface}, \citep{qian2019softtriple}), and parameter settings may improve the performance of an algorithm, leading to unfair comparison. This way, to be certain that the proposed solution actually contributes for improvements in performance, our empirical framework was carefully designed in order to keep constant as many factors as possible with a recent re-id baseline \citep{strong_baseline}\footnote{The only difference between the experiments performed on the baseline and our method was about the time to stop learning. We trained the baseline for $2000$ epochs and observed that after epoch $1120$, the  model is over-fitted and the accuracy decreases. So, we reported the best possible results for the baseline.}. This baseline has advanced the state-of-the-art performance with respect to several techniques such as \citep{kalayeh2018human},\citep{zhong2018camstyle}, and \citep{li2018harmonious}.  In summary, it is a holistic deep learning-based framework that uses a bag of tricks that are known to be particularly effective for the person re-id problem. Authors employ the ResNet-50 model as the backbone feature extractor, (fine-tuned from the ImageNet parameters) and provide a freely-available implementation in an open-source library (open-ReID\footnote{https://cysu.github.io/open-reid/notes/overview.html}). 

\begin{table*}[t]
\small
\centering
\caption{Results comparison between the baseline (top row) and our solutions for defining receptive fields, particularly tuned for the \emph{upper body} and \emph{full body}, on the RAP benchmark. mAP and Ranks $1$, $5$, and $10$ are given, for the \emph{softmax} and \emph{triplet-softmax} samplers. The best results per performance measure appear in bold.}
\label{table: results_on_RAP}
\begin{tabular}{|m{3.5cm}|M{1.35cm}|M{1.35cm}|M{1.35cm}|M{1.35cm}|M{1.35cm}|M{1.35cm}|M{1.35cm}|M{1.35cm}|} \cline{2-9}
    \multicolumn{1}{c|}{} &  \multicolumn{4}{c|}{softmax sampler} &  \multicolumn{4}{c|}{triplet-softmax sampler} \\\cline{2-9}
    \multicolumn{1}{c|}{}                        & rank=1              & rank=5               & rank=10           & mAP            & rank=1            & rank=5            & rank=10             & mAP              \\\hline
     Luo \emph{et} al.~\citep{strong_baseline}$^{*}$    & 64.1             & 81.5              & 86.8           & \textbf{45.8}           & 66.1           & 81.9           & 86.3             & 45.9             \\\hline
    Ours (Upper Body)$^{**}$                            & {62.9}             & {79.8}              & {84.7}           & {40.2}            & {64.8}           & {81.4}           & {85.9}             & {42.4}             \\\hline
    Ours (Full Body)$^{**}$                             & \textbf{65.7}    & \textbf{82.2}     & \textbf{87.2}  & 45.0   & \textbf{69.0}  & \textbf{83.6}  & \textbf{88.1}    & \textbf{46.3}    \\\hline
\multicolumn{9}{l}{$^{*}$\footnotesize{The \textbf{best possible results} occurred for \emph{triplet-softmax} sampler in epoch $1120$.}} \\ 
\multicolumn{9}{l}{$^{**}$\footnotesize{The results after $280$ epochs, lasted around 20 hours.}} \\
\end{tabular}
\vspace{-0.6 cm}
\end{table*}

\begin{table*}[b]
\small
\centering
\caption{Ablation results of the proposed attentional augmentation technique for upper-body and full-body attention types. Red and blue colors, respectively, denote the best and second-best results.}
\label{table:AblationStudy}
\begin{tabular}{|M{3cm}|M{2.5cm}|M{1.4cm}|M{1.4cm}|M{1.4cm}|M{1.4cm}|M{1.4cm}|} \hline
Attention Type & Augmentation Probability & rank=1        & rank=5         & rank=10        & rank=50         & mAP \\\hline\hline

 \multirow{5}{*}{Upper-body}   
 & 0.1       & 53.4           & 72.3           & 78.9           & 90.6            &  34.8  \\\cline{2-7}
& 0.3       & \textcolor{blue}{63.1}           & \textcolor{blue}{79.8}           & \textcolor{blue}{84.8}           & \textcolor{red}{93.2}    &  \textcolor{red}{41.1 }  \\\cline{2-7}
& 0.5       & \textcolor{red}{64.4}  & \textcolor{red}{80.3 } & \textcolor{red}{85.1 } & \textcolor{blue}{92.6}            &  \textcolor{blue}{40.2}   \\\cline{2-7}
& 0.7       & 62.1           & 78.3           & 83.0           & 91.6            &  37.7   \\\cline{2-7}
& 0.9       & 59.0           & 75.3           & 80.6           & 90.2            &  34.8   \\\hline\hline

 \multirow{2}{*}{Full-body}   
 & 0.3       & \textcolor{red}{69.0 }          & \textcolor{red}{83.6}          & \textcolor{red}{88.1}           & \textcolor{red}{94.8}            &  \textcolor{red}{46.3}  \\\cline{2-7}
& 0.5       & \textcolor{blue}{68.0}           & \textcolor{blue}{82.6}           & \textcolor{blue}{87.0}           & \textcolor{blue}{94.3}           &  \textcolor{blue}{44.6}   \\\hline

\end{tabular}
\end{table*}

\subsection{Re-ID Results}
\label{subsec:ReID_Performance}

Following the settings suggested in \citep{strong_baseline}, we re-evaluated the baseline method on the RAP dataset, using the state-of-the-art tricks such as warm-up learning rate \citep{fan2019spherereid}, random erasing data augmentation \citep{zhong2020random}, label smoothing \citep{zheng2018discriminatively}, last stride \citep{sun2018beyond}, and BNNeck \citep{strong_baseline}, alongside the conventional data augmentation transformations (i.e., random horizontal flip, random crop, and $10$-pixel-padding and original-size-crop).

Table~\ref{table: results_on_RAP} provides the overall performances based on the mean Average Precision (mAP) metric and Cumulative Match Characteristic (CMC) for ranks $1$, $5$, and $10$, denoting the possibility of retrieving at least one true positive in the top-$1$, $5$, and $10$ ranks. We evaluated both models using two sampling methods and observed a slight improvement in the performance of both methods when using the \emph{triplet-softmax} over \emph{softmax} sampler. As previously mentioned, our method could be treated as an augmentation method that requires a paired-process (i.e., exchanging the foreground and background of each pair of images), imposing a computational cost to the \textit{learning phase}. Moreover, due to increasing the learning samples from $n$ to less than $n^2$, the network needs more time and the number of epochs to converge. Therefore, learning our method (using \emph{triplet-softmax} sampler) for $280$ epochs lasted around $20$ hours with loss value $1.3$, while the baseline method accomplished $2000$ epochs after $37$ hours of learning with loss value $1.0$.

Comparison of the first and second rows of Table~\ref{table: results_on_RAP} shows that our technique with an attention on the human upper-body achieves competitive results, with less than $1.5\%$ difference with the baseline. However, the higher the rank, the closer the performance values are, such that for \emph{triplet-softmax} sampler, the difference between the rank $10$ accuracy of the baseline and our upper-body model is $0.4\%$. 

The third row of Table~\ref{table: results_on_RAP} provides the performance of the proposed method with an attention on the human full-body and --not surprisingly-- indicates that concentration on the full-body (rather than upper-body) yields more useful features for short-time person re-id. However, comparing the three rows of the result table together, we could perceive how much is the lower-body important --as a body-part with most background region? For example, when using full-body region (over the upper-body) with \emph{triplet-softmax} sampler, the rank $1$ accuracy improves $4.2\%$, while the accuracy difference of rank $1$ between the holistic baseline and full-body method is $2.9\%$, indicating that $1.3$ of our improvement (in rank $1$) over the baseline is because of attention on the lower-body and the rest ($1.6\%$) is due to focusing on the upper-body.

During the learning phase, each synthesized sample is generated with a probability between $[0, 1]$, with $0$ meaning that no changes will be done in the dataset (i.e., we use the original samples) and $1$ indicates that all samples will be transformed (augmented). We studied the effectiveness of our method for different probabilities (from $0.1$ to $0.9$) and gave the obtained results in Table~\ref{table:AblationStudy}. Overall, the optimal performance of the proposed technique is attained when the augmentation probability lies in the [$0.3, 0.5$] interval. This leads us to conclude that such intermediate probabilities of augmentation keep the discriminating information of the original data while also guarantee the transformation of enough data for yielding an effective attention mechanism.

\subsubsection{Why Does Our Method Boost The Performance?}
\label{subsec:Why_Our_Method_Boost_The_Performance?}
Our method's effectiveness could have different justifications: at first, it is known that Convolutional Neural Networks (CNNs) are heavily data-driven, and often when the more the data is available, the higher the performance will be attained. In fact, more data assist the network in exploring and exploiting the problem space's regions better, which eventuate into superior performance. Each image of the dataset can be considered as an extra point in the feature space, which augments the density of learning set. Secondly, in a supervised and label-based fashion, data augmentation is effective only when the extracted features from the synthetic data are compatible with the ground truth labels. For example, if a subject image is labeled according to the body-figure, stretching the images will make the network learn a thin person as an over-weighted person. Therefore, random erasing and random crops could provide unsafe augmentation, which does not happen in our method. Furthermore, our experiments showed that when we add human-attribute-based constraints, we generate even more realistic samples. For instance, if considering the \emph{age} constrain, to generate a new sample, we combine the $\bm{ROI}_i$ with $\bm{UR}_j$ if and only if the \emph{i}$^{th}$ and \emph{j}$^{th}$ samples have the same age label.

\section{Conclusions}
\label{sec:Conclusions_and _Future_Work}
CNNs are known to be able to autonomously find the critical regions of the input data and discriminate between foreground-background regions. However, to accomplish such a challenging goal, they demand large volumes of learning data, which can be hard to collect and particularly costly to annotate, in case of supervised learning problems. In this paper, we described a solution based on data segmentation and swapping, that interchanges segments \emph{apriori} deemed to be important/irrelevant for the network responses. The proposed method can be seen as a data augmentation solution that implicitly also empowers the network to improve its \emph{receptive fields inference skills}. In practice terms, during the learning phase, we provide the network with an attentional mechanism derived from prior information (i.e., annotations and body masks), that not only determines the critical regions of the input data but also provides important cues about any useless input segments that should be disregarded from the decision process. Finally, it is important to stress that, in \emph{test} time, samples are provided without any segmentation mask, which lowers the computational burden with respect to previously proposed explicit attention mechanisms. As a proof-of-concept, our experiments were carried out in the highly challenging pedestrian re-identification problem, and the results show that our approach --as a complementary data augmentation technique-- could contributes to significant improvements in performance of the state-of-the-art re-id baseline.


\section*{Acknowledgments}
\small
This research is funded by the "FEDER, Fundo de Coesao e Fundo Social Europeu" under the "PT2020 - Portugal 2020" program, "IT: Instituto de Telecomunicações" and "TOMI: City's Best Friend." Also, the work is funded by FCT/MEC through national funds and, when applicable, co-funded by the FEDER PT2020 partnership agreement under the project UID/EEA/50008/2019.


\end{document}